\documentclass{article} 
\usepackage{iclr2027_conference,times}

\usepackage{amsmath,amsfonts,bm}

\def\eqref#1{equation~\ref{#1}}

\def\1{\bm{1}}

\DeclareMathAlphabet{\mathsfit}{\encodingdefault}{\sfdefault}{m}{sl}
\SetMathAlphabet{\mathsfit}{bold}{\encodingdefault}{\sfdefault}{bx}{n}

\newcommand{\best}[1]{\cellcolor{red!40}#1}
\newcommand{\second}[1]{\cellcolor{orange!40}#1}
\newcommand{\third}[1]{\cellcolor{yellow!40}#1}
\usepackage{hyperref}
\usepackage{url}
\usepackage{capt-of}
\usepackage{graphicx}
\usepackage{wrapfig}
\usepackage{xcolor}
\usepackage{multirow}
\usepackage{booktabs}
\usepackage{caption}
\definecolor{separatorgray}{RGB}{200,200,200}
\usepackage[table]{xcolor}
\definecolor{darkgreen}{RGB}{0,170,0}
\definecolor{oursbg}{RGB}{255,247,247}
\title{PixelSR: Pixel-Space Orthogonal Flow Decoupling for Real-World Image Super-Resolution}

\author{Antiquus S.~Hippocampus, Natalia Cerebro \& Amelie P. Amygdale \thanks{ Use footnote for providing further information
about author (webpage, alternative address)---\emph{not} for acknowledging
funding agencies.  Funding acknowledgements go at the end of the paper.} \\
Department of Computer Science\\
Cranberry-Lemon University\\
Pittsburgh, PA 15213, USA \\
\texttt{\{hippo,brain,jen\}@cs.cranberry-lemon.edu} \\
\And
Ji Q. Ren \& Yevgeny LeNet \\
Department of Computational Neuroscience \\
University of the Witwatersrand \\
Joburg, South Africa \\
\texttt{\{robot,net\}@wits.ac.za} \\
\AND
Coauthor \\
Affiliation \\
Address \\
\texttt{email}
}

\makeatletter
\def\makevruler[#1][#2][#3][#4][#5]{}
\makeatother

\begin{document}

\maketitle

\vspace{-0.6cm}
\begin{center}
    \includegraphics[width=\textwidth]{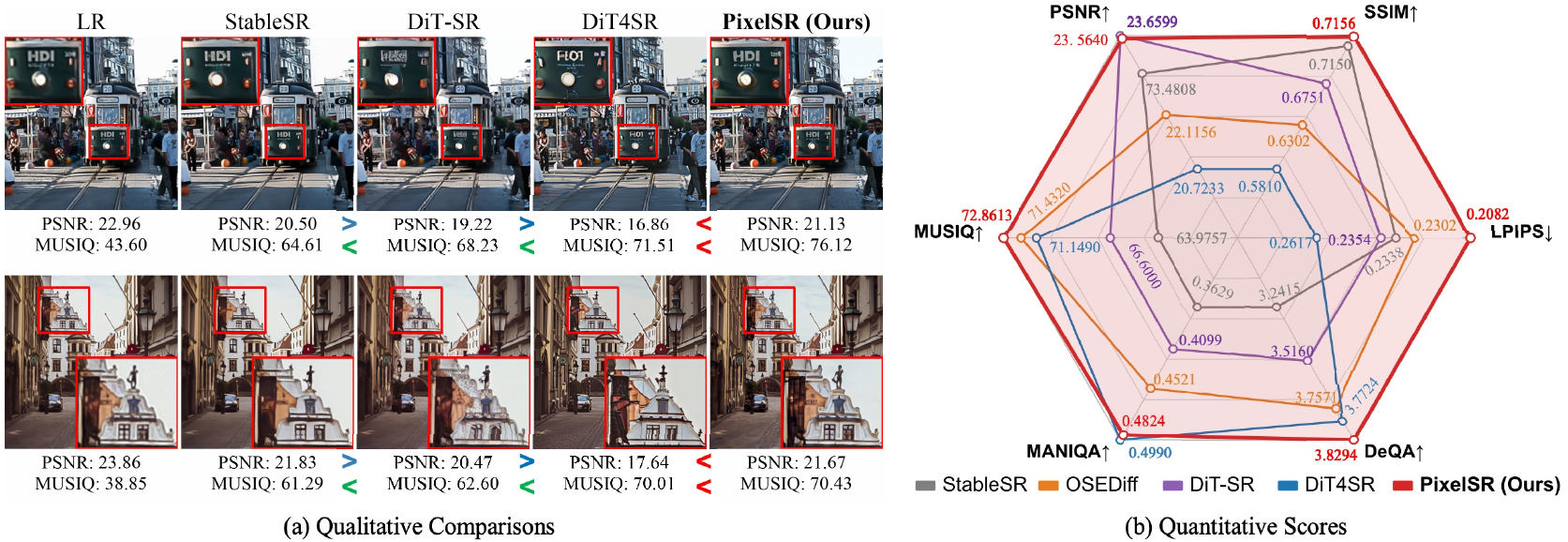}
    \vspace{-0.2cm}
    \captionof{figure}{
\textbf{Fidelity--realism \textit{seesaw effect} on ScreenSR.}
(a) Qualitatively, existing latent-space methods sacrifice structural fidelity for realism, leading to text distortions and hallucinations (see crops). (b) Quantitatively, higher perceptual quality (MANIQA↑, MUSIQ↑, DeQA↑) typically correlates with degraded fidelity (PSNR↑, SSIM↑, LPIPS↓). In contrast, our PixelSR successfully reconciles this trade-off, achieving superior performance in both dimensions.
    }
    \label{fig:teaser}
\end{center}

\begin{abstract}

Latent diffusion models have recently achieved remarkable progress in Real-world Image Super-Resolution (Real-SR). 
However, existing methods consistently exhibit a persistent \textit{seesaw effect}: enhancing perceptual realism compromises structural fidelity. 
This effect, we argue, is not intrinsic to diffusion models, but largely stems from the architectural constraints of prevailing two-stage frameworks. 
To address this challenge, we propose \textbf{PixelSR}, a novel single-stage Pixel Diffusion Transformer for Real-SR that operates directly in pixel space.
Specifically,  PixelSR consists of two core designs: Pixel-space Conditional Diffusion and Orthogonal Flow Decoupling.  
The former establishes an effective pixel-space restoration framework by integrating spatially aligned LR conditioning, visual semantic guidance, and lightweight texture refinement, thereby bypassing VAE compression and enabling end-to-end optimization directly in RGB space.
Complementarily, the latter decomposes the asymmetric flow target into a low-rank semantic component and a noise-free orthogonal texture component. By routing these components to specialized branches, it achieves decoupled optimization with reduced mutual interference.
Benefiting from these designs, our PixelSR restores perceptually realistic details while faithfully preserving structural content.
Extensive experiments across multiple benchmarks demonstrate that PixelSR not only achieves competitive performance, but also effectively mitigates the fidelity--realism \textit{seesaw effect}.

\end{abstract}

\section{Introduction}
Real-world Image Super-Resolution (Real-SR) aims to recover High-Resolution (HR) images from Low-Resolution (LR) observations undergoing complex and unknown degradations \citep{ODTSR,VPD-SR}. 
Recently, Latent Diffusion Models (LDMs) have dominated this task by performing denoising in the compressed latent space of a pre-trained VAE, substantially reducing computational and memory costs \citep{LDM,RealOSR,Seesr}. 

Despite this efficiency, prior studies have highlighted fidelity limitations of latent representations, with text and high-frequency details susceptible to degradation even during VAE reconstruction alone~\citep{lightingdit,medi12026decoupling,SD3,gupta2024photorealistic}.
These limitations are particularly relevant to Real-SR, where 
LDMs typically exhibit a \textit{seesaw effect}: improving perceptual realism often comes at the cost of fidelity to the LR, as illustrated in Figure~\ref{fig:teaser}.
Qualitatively, this effect manifests as structural distortions, severe text degradation, and content hallucinations.
Quantitatively, improvements in no-reference perceptual quality, as measured by MANIQA and MUSIQ, are consistently accompanied by lower PSNR and SSIM, indicating degraded reference-based fidelity.

We argue that this \textit{seesaw effect} is not an insurmountable limit of diffusion models, but rather stems from two structural bottlenecks of the LDM paradigm: 
(1) pre-trained VAEs inevitably perform lossy reconstruction, discarding high-frequency details and fine-grained structures, which inherently limits the upper bound of structural fidelity; 
(2) the disjointed two-stage pipeline prevents end-to-end optimization, causing reconstruction and approximation errors to cascade and accumulate between latent generation and pixel decoding. 
These architectural limitations motivate a fundamental question: \textit{Can we bypass the latent bottleneck and perform Real-SR directly in the pixel space?}

Emerging advancements in pixel-space diffusion have demonstrated the feasibility of generative modeling directly on raw pixels without relying on a pre-trained VAE.
However, bypassing the VAE bottleneck alone does not automatically resolve the persistent fidelity--realism trade-off. In conventional flow matching, predicting the standard velocity target $\boldsymbol{v} := \boldsymbol{\epsilon} - \boldsymbol{x}_0$ couples full-dimensional noise  with clean-image reconstruction, requiring a single network to simultaneously perform high-level semantic generation and low-level texture recovery~\citep{AsymFlow}.
This task entanglement is particularly problematic for Real-SR: the former role relies on generative priors to synthesize realistic content, whereas the latter demands strict pixel-level alignment with the LR observation. To solve this challenge, several methods attempt to ease this optimization burden through architectural bypasses or direct clean-image prediction~\citep{simplerdiffusion,JiT}.
Nevertheless, they either complicate the transformer recipe, suffer from numerical ill-conditioning, or fail to explicitly partition these conflicting roles within the denoising flow. Therefore, the key challenge of pixel-space Real-SR lies in how to orthogonally decouple these two roles within a unified, conditional framework, thereby reconciling the tension between perceptual realism and structural fidelity.

Based on the above analysis, we propose \textbf{PixelSR}, a novel single-stage pixel-space diffusion framework for Real-SR that aims to reconcile structural fidelity and perceptual realism, as displayed in Figure~\ref{fig:framework}.  
Specifically,  our PixelSR is built upon Pixel-space Conditional Diffusion (PCD), which integrates spatially aligned LR conditioning, visual semantic guidance, and lightweight texture refinement to perform end-to-end restoration directly in RGB space, thereby eliminating the need for lossy VAE encoding and decoding.
To alleviate task entanglement in conventional flow matching, we further develop Orthogonal Flow Decoupling (OFD) by decomposing the asymmetric velocity $\boldsymbol{u}_{\mathrm{A}}:=\boldsymbol{P}\boldsymbol{\epsilon}-\boldsymbol{x}_0$ into a low-rank semantic component $\boldsymbol{P}\boldsymbol{u}_{\mathrm{A}}$ and an orthogonal texture component $(\boldsymbol{I}-\boldsymbol{P})\boldsymbol{u}_{\mathrm{A}}$, enabling specialized optimization with reduced interference. 



    


Our main contributions are summarized as follows:
\begin{itemize}
\item  We present \textbf{PixelSR}, a single-stage end-to-end Pixel Diffusion Transformer for Real-SR. Its PCD framework adapts the original diffusion backbone to operate directly in pixel space, bypassing the lossy VAE bottleneck.
\item We propose OFD to mitigate task entanglement in  pixel-space flow matching. By decomposing  asymmetric velocity target into a low-rank semantic  and an orthogonal texture component,  OFD assigns them to specialized modules to minimize optimization interference.
\item Extensive experiments across multiple Real-SR benchmarks demonstrate that PixelSR delivers consistently strong performance across both fidelity- and perception-oriented metrics, effectively mitigating the fidelity--realism \textit{seesaw effect}.
\end{itemize}

\vspace{-0.4cm}
\section{Related Work}

\subsection{Real-World Image Super-Resolution}

The methodology of Real-SR has evolved from fidelity-oriented reconstruction to perceptual detail generation. 
Early deep-learning approaches, including convolutional neural networks and vision Transformers~\citep{srcnn,swinir}, primarily optimize pixel-wise reconstruction objectives (e.g., $L_1$ or $L_2$ losses).
Although these methods preserve structural integrity and achieve favorable distortion metrics, they inherently suffer from over-smoothing due to the regression-to-the-mean problem. 
To improve perceptual quality, generative adversarial networks~\citep{Real-esrgan,bsrgan} are introduced by utilizing adversarial and perceptual losses to recover high-frequency textures under complex degradations. 
However, despite producing sharper details, they frequently produce unnatural visual artifacts.

More recently, diffusion models have emerged as a powerful alternative for realistic detail generation. While early approaches train conditional diffusion models specifically for SR~\citep{sr3,resshift}, recent methods increasingly adapt large-scale pre-trained text-to-image priors~\citep{StableSR,DiT4SR}. Despite their impressive perceptual quality, most of the latter operate in a VAE-compressed latent space, potentially losing fine-grained structures and separating latent prediction from final RGB reconstruction. 
To improve deployment efficiency, recent studies attempt to bypass or streamline the VAE stage via distillation~\citep{GenDRPix,ADC}. 
However, as post-hoc compression, they still fail to achieve genuine, pixel-space joint optimization. In contrast, our PixelSR establishes pixel-space flow modeling, seamlessly integrating restoration and reconstruction via explicit semantic–texture decoupling.

\subsection{Pixel-Space Diffusion Models}
Early diffusion models operate directly in pixel space, typically using U-Net architectures with multi-scale features and skip connections to propagate high-dimensional signals~\citep{ddpm}. 
Subsequently, LDMs shift the diffusion process to VAE-compressed latent representations, substantially reducing computational and memory costs and thereby enabling more scalable training and generation~\citep{LDM,flowmatching}. 

Lately, however, the paradigm is shifting back toward pixel-space modeling to construct self-contained, tokenizer-free generative frameworks.
Specifically, these studies mainly address the high-dimensional prediction burden through architectural design or prediction-objective reformulation. For example, PixelDiT~\citep{Pixeldit} and DiP~\citep{DiP} augment Transformer backbones with pixel-level or U-Net-style decoders to facilitate fine-grained reconstruction.
Alternatively, JiT~\citep{JiT} adopts clean-image prediction ($\boldsymbol{x}_0$-prediction) under the low-dimensional manifold assumption, while AsymFlow~\citep{AsymFlow} restricts noise modeling to a low-rank subspace via an asymmetric velocity target $\boldsymbol{u}_{\mathrm{A}}$.  Nevertheless, the monolithic targets of these methods still couple semantic generation with texture reconstruction in the same prediction space. Unlike these methods, our OFD explicitly factorizes prediction objectives into a low-rank semantic component and a noise-free orthogonal texture component for specialized optimization.

\begin{figure*}[t]
    \centering
    \includegraphics[width=\linewidth]{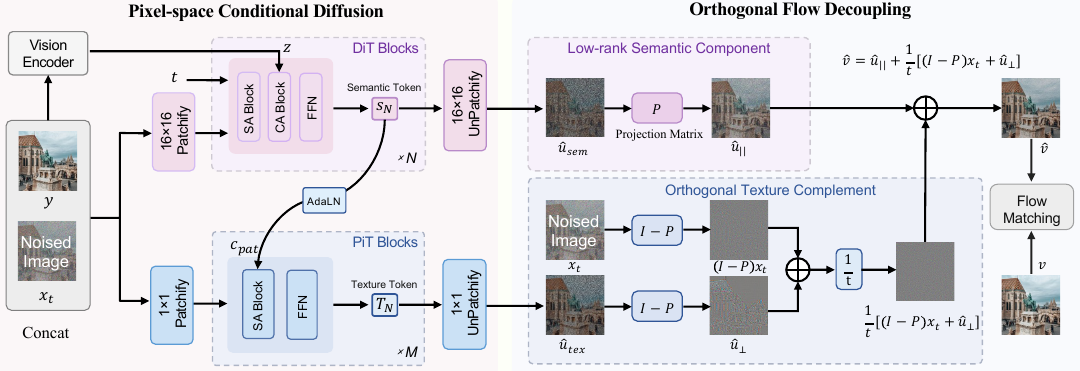}
    \vspace{-0.2cm}
\caption{
\textbf{Overview of the proposed PixelSR framework.}
PixelSR comprises Pixel-space Conditional Diffusion (PCD) and Orthogonal Flow Decoupling (OFD). 
\textbf{Left:} PCD performs end-to-end RGB-space restoration using spatially aligned LR conditioning, visual semantic guidance, and lightweight pixel-level texture refinement. 
\textbf{Right:} OFD factorizes the asymmetric velocity target into a projected low-rank component and a deterministic orthogonal residual, which empirically exhibit semantic- and texture-dominant characteristics, respectively. Their predictions are assigned to specialized DiT and PiT branches and recombined into the full velocity for flow matching, enabling task-aligned optimization with reduced interference.
}
    \label{fig:framework}
    \vspace{-0.2cm}
\end{figure*}

\section{Methodology}
\label{sec:method}



\subsection{Preliminaries}
\label{sec:preliminaries}

\paragraph{Conditional Flow Matching.}
Let $\boldsymbol{x}_0$ denote a target HR image in pixel space and $\boldsymbol{y}$ its spatially aligned LR observation. Following Rectified Flow~\citep{flowmatching}, we adopt a linear probability path from the data distribution at $t=0$ to a standard Gaussian distribution at $t=1$:
\begin{equation}
\boldsymbol{x}_t
=
(1-t)\boldsymbol{x}_0+t\boldsymbol{\epsilon},
\qquad
\boldsymbol{\epsilon}\sim\mathcal{N}(\boldsymbol{0},\boldsymbol{I}),
\quad t\in[0,1].
\label{eq:linear_path}
\end{equation}
The associated velocity is:
\begin{equation}
\boldsymbol{v}
=
\frac{\mathrm{d}\boldsymbol{x}_t}{\mathrm{d}t}
=
\boldsymbol{\epsilon}-\boldsymbol{x}_0.
\label{eq:standard_velocity}
\end{equation}
Given a velocity estimate
$\widehat{\boldsymbol{v}}_{\theta}(\boldsymbol{x}_t,t,\boldsymbol{y})$,
Conditional Flow Matching minimizes:
\begin{equation}
\mathcal{L}_{\mathrm{CFM}}
=
\mathbb{E}_{t,\boldsymbol{x}_0,\boldsymbol{y},\boldsymbol{\epsilon}}
\left[
\left\|
\widehat{\boldsymbol{v}}_{\theta}
-
\boldsymbol{v}
\right\|_2^2
\right].
\label{eq:cfm_loss}
\end{equation}
At inference, the learned velocity field is integrated from $t=1$ to $t=0$, transforming Gaussian noise into an HR image conditioned on $\boldsymbol{y}$.

\paragraph{Prediction Parameterizations.}
The flow matching loss and the direct network output need not lie in the same space. Under standard $\boldsymbol{v}$-prediction, the network directly estimates:
\begin{equation}
\widehat{\boldsymbol{v}}_{\theta}
=
\operatorname{Net}_{\theta}
(\boldsymbol{x}_t,t,\boldsymbol{y}).
\end{equation}
However, predicting $\boldsymbol{v}=\boldsymbol{\epsilon}-\boldsymbol{x}_0$ requires the network to preserve full-dimensional noise, which is challenging in pixel space. Motivated by the low-dimensional manifold hypothesis, $\boldsymbol{x}_0$-prediction instead  outputs~\citep{JiT}:
\begin{equation}
\widehat{\boldsymbol{x}}_0
=
\operatorname{Net}_{\theta}
(\boldsymbol{x}_t,t,\boldsymbol{y}),
\qquad
\widehat{\boldsymbol{v}}_{\theta}
=
\frac{\boldsymbol{x}_t-\widehat{\boldsymbol{x}}_0}{t}.
\label{eq:x0_prediction}
\end{equation}
Although this avoids directly predicting full-dimensional noise, computing the flow matching loss in velocity space requires converting $\widehat{\boldsymbol{x}}_0$ into $\widehat{\boldsymbol{v}}$, which becomes numerically ill-conditioned as $t$ approaches zero.

Consequently, AsymFlow~\citep{AsymFlow} provides an intermediate parameterization by restricting the noise term to a low-rank subspace. For each $D$-dimensional patch token, let
$\boldsymbol{A}\in\mathbb{R}^{D\times r}$, $r<D$, satisfy
$\boldsymbol{A}^{\mathsf T}\boldsymbol{A}=\boldsymbol{I}_r$.
The corresponding orthogonal projector is defined as:
\begin{equation}
\boldsymbol{P}
=
\boldsymbol{A}\boldsymbol{A}^{\mathsf T},
\qquad
\boldsymbol{P}^{2}=\boldsymbol{P},
\label{eq:projector}
\end{equation}
where \(\boldsymbol{P}\) is shared across  all patch tokens.  AsymFlow defines the asymmetric velocity target as:
\begin{equation}
\boldsymbol{u}_{\mathrm{A}}
:=
\boldsymbol{P}\boldsymbol{\epsilon}
-
\boldsymbol{x}_0.
\label{eq:asymmetric_velocity}
\end{equation}
It recovers standard $\boldsymbol{v}$-prediction when
$\boldsymbol{P}=\boldsymbol{I}$ and clean-image prediction, up to a sign, when $\boldsymbol{P}=\boldsymbol{0}$. Therefore, intermediate ranks  balance low-rank noise prediction with full-dimensional clean-image prediction. Nevertheless, $\boldsymbol{u}_{\mathrm{A}}$ remains a monolithic network output in AsymFlow. 

To address this limitation, we first establish Pixel-space Conditional Diffusion (PCD) as an end-to-end RGB 
restoration framework. Subsequently, we introduce Orthogonal Flow Decoupling (OFD) to factorize $\boldsymbol{u}_{\mathrm{A}}$ into complementary objectives for task-specialized optimization.

\subsection{Pixel-space Conditional Diffusion}
\label{sec:pcd}
As the restoration backbone of PixelSR, PCD establishes a single-stage Pixel Diffusion Transformer for Real-SR. It adopts an augmented DiT backbone~\citep{lightingdit} and directly instantiates the conditional flow through $\boldsymbol{x}_0$-prediction in RGB space. To adapt the original generative backbone to Real-SR, we incorporate three complementary designs: spatially aligned LR conditioning, visual semantic guidance, and lightweight pixel-level texture refinement, as shown in Figure~\ref{fig:framework}.

\paragraph{Spatially Aligned LR Conditioning.}
Given an LR observation $\boldsymbol{y}$, we first resize it to the target resolution determined by the upscaling factor, obtaining $\widetilde{\boldsymbol{y}}\in\mathbb{R}^{B\times C\times H\times W}$. Then, we construct the noisy state $\boldsymbol{x}_t\in\mathbb{R}^{B\times C\times H\times W}$ on the same spatial grid and concatenate them channel-wise:
\begin{equation}
\boldsymbol{h}_t
=
\operatorname{Concat}
\left(
\boldsymbol{x}_t,\widetilde{\boldsymbol{y}}
\right)
\in
\mathbb{R}^{B\times 2C\times H\times W}.
\label{eq:pcd_input}
\end{equation}
This pixel-aligned conditioning provides each noisy location with its corresponding LR evidence, imposing dense structural constraints throughout restoration. Subsequently, the resulting input is divided into non-overlapping $p\times p$ patches and embedded into DiT tokens:
\begin{equation}
\boldsymbol{f}_0
=
\operatorname{PatchEmbed}_{p}(\boldsymbol{h}_t)
\in
\mathbb{R}^{B\times L\times D_{\mathrm{DiT}}},
\label{eq:pcd_patch_tokens}
\end{equation}
where $D_{\mathrm{DiT}}$ denotes the hidden dimension of the DiT backbone and $L=(H/p)(W/p)$ is the number of patch tokens.

\paragraph{Visual Semantic Guidance.}
Spatially aligned LR pixels may provide insufficient semantic evidence under severe degradation. We therefore use a frozen pre-trained vision encoder to extract high-level representations:
\begin{equation}
\boldsymbol{z}
=
\Pi_{\mathrm{vis}}
\left(
E_{\mathrm{vis}}(\boldsymbol{y})
\right),
\label{eq:visual_semantics}
\end{equation}
where $\Pi_{\mathrm{vis}}$ projects the visual features into the DiT conditioning space.  We further inject $\boldsymbol{z}$ into each DiT block through cross-attention:
\begin{equation}
\boldsymbol{f}_{n+1}
=
\operatorname{DiT}_{n}
\left(
\boldsymbol{f}_{n},t,\boldsymbol{z}
\right),
\qquad n=0,\ldots,N-1.
\label{eq:dit_blocks}
\end{equation}
Accordingly, the concatenated LR input anchors spatial structures, while the visual embeddings provide complementary semantic guidance for severely degraded content.

\paragraph{Lightweight Texture Refinement.}

To complement patch-level semantics with fine-grained textures, we introduce a lightweight Pixel Decoder composed of PiT blocks~\citep{Pixeldit}. It operates on pixel-level tokens under the global guidance of the DiT backbone. Specifically, each $p\times p$ region of $\boldsymbol{h}_t$ is represented as $p^2$ pixel tokens:
\begin{equation}
\boldsymbol{q}_0
=
\operatorname{PixelEmbed}_{p}(\boldsymbol{h}_t)
\in
\mathbb{R}^{(BL)\times p^2\times D_{\mathrm{PiT}}},
\label{eq:pixel_tokens}
\end{equation}
where $D_{\mathrm{PiT}}$ denotes the hidden dimension of the lightweight Pixel Decoder.
After that, the final DiT features $\boldsymbol{f}_{N}$ are expanded into patch-wise conditions $\boldsymbol{c}_{\mathrm{pat}}$ and injected into the corresponding PiT blocks through AdaLN:
\begin{equation}
\boldsymbol{q}_{m+1}
=
\operatorname{PiT}_{m}
\left(
\boldsymbol{q}_{m},
\boldsymbol{c}_{\mathrm{pat}}
\right),
\qquad m=0,\ldots,M-1.
\label{eq:pit_blocks}
\end{equation}
This design further refines local textures under the global structural and semantic context provided by the DiT backbone.


Overall, the basic PCD predicts the complete clean image as
$\widehat{\boldsymbol{x}}_0=
\operatorname{PixelUnpatchify}(\Phi_{\mathrm{pix}}(\boldsymbol{q}_M))$
and optimizes it via Eqs.~\ref{eq:cfm_loss} and \ref{eq:x0_prediction}, establishing an effective framework for direct pixel-space restoration. However, its monolithic $\boldsymbol{x}_0$ target still couples semantic generation with texture reconstruction in the same prediction space. To this end, we next introduce OFD to explicitly decouple these heterogeneous objectives.

\subsection{Orthogonal Flow Decoupling}
\label{sec:ofd}

As illustrated on the right of Figure~\ref{fig:framework}, our OFD builds upon the asymmetric velocity target but goes beyond its monolithic prediction by explicitly factorizing and assigning its complementary components to task-specialized branches.

\paragraph{Orthogonal Target Factorization.}
Unlike AsymFlow, which predicts $\boldsymbol{u}_{\mathrm{A}}$ as a unified target,  OFD explicitly exposes its complementary projections. Using the projector properties in Eq.~\ref{eq:projector}, we obtain:
\begin{equation}
\boldsymbol{P}\boldsymbol{u}_{\mathrm{A}}
=
\boldsymbol{P}\boldsymbol{v},
\quad
(\boldsymbol{I}-\boldsymbol{P})\boldsymbol{u}_{\mathrm{A}}
=
-(\boldsymbol{I}-\boldsymbol{P})\boldsymbol{x}_0.
\label{eq:ofd_decomposition}
\end{equation}
The projected component $\boldsymbol{P}\boldsymbol{u}_{\mathrm{A}}$ retains stochastic velocity prediction within the low-rank subspace, whereas its orthogonal complement $(\boldsymbol{I}-\boldsymbol{P})\boldsymbol{u}_{\mathrm{A}}$ is completely noise-free, reducing the objective in this subspace to a deterministic reconstruction of the clean-image residual. Moreover, these two components lie in mutually orthogonal subspaces, providing non-overlapping prediction objectives.

\begin{wrapfigure}{r}{0.475\textwidth}
\vspace{-0.4cm}
    \centering
    \includegraphics[width=\linewidth]{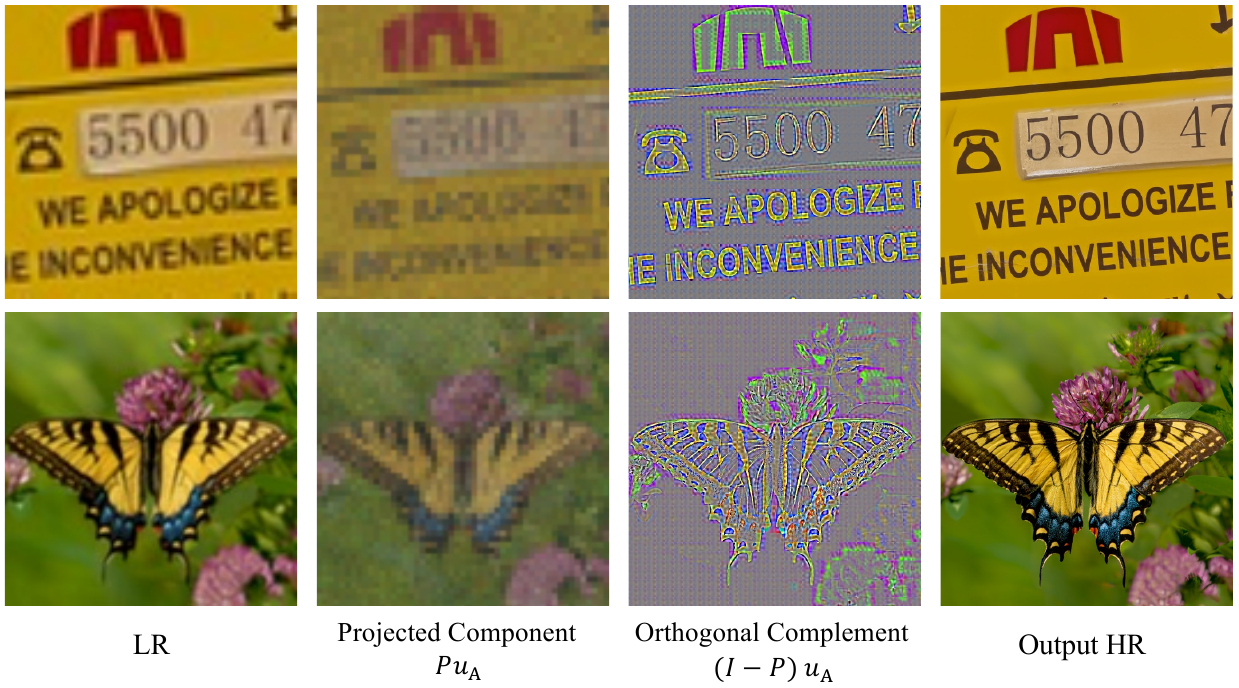}
    \vspace{-0.7cm}
    \caption{\textbf{Visualization of orthogonal target factorization.}
    With the Procrustes-aligned projector, the projected component
    mainly retains global content and coarse structures, while its
    orthogonal complement emphasizes edges, text patterns, and
    high-frequency textures.}
    \label{fig:visualization}
    \vspace{-0.3cm}
\end{wrapfigure}
\paragraph{Semantic--Texture Interpretation.}
Although this factorization holds for any orthogonal projector, the functional roles of its subspaces depend on the projection basis $\boldsymbol{P}$. We explore three constructions: a jointly learned basis, an SVD basis derived from pixel statistics, and a Procrustes basis aligned with pre-trained VAE latent tokens. The Procrustes basis performs best empirically, indicating that the compact semantic organization of VAE representations provides a useful prior for defining the low-rank subspace. 
Crucially, the VAE is used only to initialize the projector offline and is absent from PixelSR training and inference, providing an alignment starting point rather than imposing its frozen reconstruction bottleneck.
Detailed derivations are provided in the \textbf{Appendix} \ref{sec:projector_construction}.
Furthermore, as visualized in Figure~\ref{fig:visualization}, the projected component mainly preserves global content and coarse structures, whereas the orthogonal component concentrates on edges, textual patterns, and high-frequency textures. Therefore, we associate $\boldsymbol{P}\boldsymbol{u}_{\mathrm{A}}$ with semantic generation and $(\boldsymbol{I}-\boldsymbol{P})\boldsymbol{u}_{\mathrm{A}}$ with texture reconstruction. Notably, the latter additionally provides noise-free supervision for recovering fine-grained details.

\vspace{-0.2cm}
\begin{table*}[t]
\centering
\scriptsize
\caption{\textbf{Quantitative comparison on LSDIR, RealSR, ScreenSR, and UGCSR.}  \colorbox{red!40}{Red} indicates the best performance, \colorbox{orange!40}{orange} indicates the second-best performance, and \colorbox{yellow!40}{yellow} indicates the third-best performance. \(\uparrow\): higher is better; \(\downarrow\): lower is better.}
\vspace{-0.3cm}
\renewcommand{\arraystretch}{1.2}
\setlength{\tabcolsep}{3pt}
\resizebox{\textwidth}{!}{
\begin{tabular}{ccccccccccccc}
\toprule
\multirow{2}{*}{Dataset} &
\multirow{2}{*}{Method} &
\multirow{2}{*}{Steps} &
\multirow{2}{*}{\shortstack{Inference\\Time (s)}} &
\multicolumn{4}{c}{Reference-based Metrics} &
\multicolumn{4}{c}{No-Reference Metrics} &
\multirow{2}{*}{Rank $\downarrow$} \\
\cmidrule(lr){5-8} \cmidrule(lr){9-12}
& & & &
PSNR$\uparrow$ & SSIM$\uparrow$ & LPIPS$\downarrow$ & FID$\downarrow$ &
NIQE$\downarrow$ & MUSIQ$\uparrow$ & MANIQA$\uparrow$ & DeQA$\uparrow$ & \\
\midrule

& StableSR & 200 & 5.75 & 20.4946 & 0.5227 & 0.3193 & 60.0564 & 4.6047 & 62.7178 & 0.3254 & 3.0055 & 9 \\
& DiffBIR & 50 & 5.08 & 20.5132 & 0.4899 & 0.2651 & 58.4458 & 4.4425 & \second{73.2548} & \best{0.6209} & \best{4.1696} & 4 \\
& DiT4SR & 40 & 5.26 & 19.2087 & 0.4804 & 0.2723 & 59.0254 & 4.2242 & 72.6805 & 0.5067 & 3.8741 & 8 \\
& VOSR & 25 & 3.86 & 20.5993 & 0.5111 & \third{0.2386} & 46.4918 & \best{3.8768} & 72.9454 & \third{0.5194} & \third{4.0873} & \second{2} \\
LSDIR & DiT-SR & 15 & 1.84 & \best{21.1127} & \best{0.5490} & 0.2457 & \third{41.7907} & 4.2975 & 69.5674 & 0.4341 & 3.8280 & 6 \\
& OSEDiff & 1 & 0.45 & 20.3920 & 0.5215 & 0.2703 & 59.5401 & \third{4.0240} & 72.3407 & 0.4586 & 3.8411 & 7 \\
& SinSR & 1 & 0.32 & \second{20.9234} & \second{0.5375} & \second{0.2268} & \best{39.6452} & 4.6581 & 69.6398 & 0.4654 & 3.9294 & \third{3} \\
& CODSR & 1 & 6.24 & 20.3326 & 0.5160 & 0.2499 & 53.3267 & \second{3.9177} & \best{73.2553} & 0.4705 & 3.8952 & 5 \\
& \textbf{PixelSR (Ours)} & 25 & 4.45 & \third{20.6907} & \third{0.5270} & \best{0.2203} & \second{40.5007} & 4.2464 & \third{73.1895} & \second{0.5661} & \second{4.0936} & \best{1} \\
\midrule

& StableSR & 200 & 5.79 & 24.3860 & 0.6777 & 0.3194 & 149.2795 & 6.2610 & 47.8669 & 0.2851 & 2.4946 & 9 \\
& DiffBIR & 50 & 5.01 & 24.1268 & 0.6315 & 0.3680 & 129.6928 & 5.7679 & 69.4416 & \best{0.5595} & \best{3.8413} & 6 \\
& DiT4SR & 40 & 5.20 & 22.9574 & 0.6537 & 0.3190 & \third{121.2180} & 5.9232 & 67.3697 & 0.4559 & 3.4006 & 8 \\
& VOSR & 25 & 3.84 & 24.8590 & 0.7024 & 0.3040 & 135.6466 & 5.6586 & \second{69.8075} & \third{0.4944} & 3.5850 & 4 \\
RealSR & DiT-SR & 15 & 1.84 & \third{25.2487} & \best{0.7328} & 0.2976 & 148.3538 & \third{5.5714} & 65.8281 & 0.4500 & 3.5583 & 5 \\
& OSEDiff & 1 & 0.48 & 24.5348 & \third{0.7166} & \third{0.2972} & 123.1172 & 5.7055 & 69.1131 & 0.4710 & \second{3.6600} & \third{3} \\
& SinSR & 1 & 0.32 & \best{25.5445} & \second{0.7217} & 0.3181 & 135.3449 & 6.2829 & 60.8381 & 0.4000 & 3.3745 & 7 \\
& CODSR & 1 & 6.37 & 24.7344 & 0.7100 & \best{0.2789} & \second{115.6569} & \best{5.2396} & \best{70.4197} & 0.4689 & \third{3.6330} & \second{2} \\
& \textbf{PixelSR (Ours)} & 25 & 4.48 & \second{25.3360} & \third{0.7166} & \second{0.2922} & \best{112.8905} & \second{5.5672} & \third{69.5941} & \second{0.5381} & 3.6116 & \best{1} \\
\midrule

& StableSR & 200 & 5.71 & \third{23.4808} & \second{0.7150} & 0.2338 & 72.4313 & 4.5287 & 63.9757 & 0.3629 & 3.2415 & 8 \\
& DiffBIR & 50 & 5.18 & 22.6359 & 0.6044 & 0.2339 & \best{60.3978} & 4.4604 & 71.6173 & \best{0.5998} & \best{4.0741} & 4 \\
& DiT4SR & 40 & 5.27 & 20.7233 & 0.5810 & 0.2617 & 78.7023 & \third{4.1735} & 71.1490 & \third{0.4990} & 3.7724 & 8 \\
& VOSR & 25 & 3.84 & 22.2853 & 0.6241 & \third{0.2167} & \third{67.5358} & 4.2239 & \third{72.4115} & \second{0.5190} & \second{4.0030} & \second{2} \\
ScreenSR & DiT-SR & 15 & 1.86 & \best{23.6599} & \third{0.6751} & 0.2354 & 71.9429 & 4.4406 & 66.6000 & 0.4099 & 3.5160 & 7 \\
& OSEDiff & 1 & 0.43 & 22.1156 & 0.6302 & 0.2302 & 74.9355 & \second{4.0782} & 71.4320 & 0.4521 & 3.7571 & 5 \\
& SinSR & 1 & 0.32 & 23.4096 & 0.6619 & 0.2225 & \second{65.8214} & 4.6827 & 66.3274 & 0.4281 & 3.6553 & 5 \\
& CODSR & 1 & 6.09 & 22.1222 & 0.6253 & \second{0.2123} & 67.5680 & \best{4.0608} & \second{72.4353} & 0.4705 & 3.8178 & \third{3} \\
& \textbf{PixelSR (Ours)} & 25 & 4.48 & \second{23.5640} & \best{0.7156} & \best{0.2082} & 69.1307 & 4.3284 & \best{72.8613} & 0.4824 & \third{3.8294} & \best{1} \\
\midrule

& StableSR & 200 & 5.71 & \second{24.3958} & \second{0.7843} & \third{0.2056} & 75.4312 & \second{4.7718} & 70.8247 & 0.4715 & 3.5726 & 4 \\
& DiffBIR & 50 & 5.19 & 23.0754 & 0.6601 & 0.2243 & \second{67.5485} & 5.7892 & 74.0110 & \best{0.6470} & \best{4.0048} & \third{3} \\
& DiT4SR & 40 & 5.17 & 21.7385 & 0.6351 & 0.2534 & 84.8692 & \best{4.6058} & 72.9437 & 0.5088 & 3.7754 & 9 \\
& VOSR & 25 & 3.86 & 21.8930 & 0.6419 & 0.2609 & 106.5576 & 5.2074 & \third{74.7371} & \third{0.5451} & \third{3.9251} & 8 \\
UGCSR & DiT-SR & 15 & 1.87 & \third{23.9624} & \third{0.7187} & 0.2271 & 76.4899 & 5.0916 & 69.9280 & 0.4558 & 3.6601 & 6 \\
& OSEDiff & 1 & 0.46 & 22.3983 & 0.6726 & 0.2314 & 86.6685 & 5.2381 & 73.5825 & 0.5089 & 3.8033 & 6 \\
& SinSR & 1 & 0.33 & 23.8187 & 0.7079 & 0.2147 & \third{68.0871} & 5.4829 & 69.2716 & 0.4590 & 3.7103 & 5 \\
& CODSR & 1 & 6.21 & 22.2322 & 0.6642 & \second{0.2019} & 73.4316 & \third{4.8300} & \second{74.9784} & 0.5416 & 3.9048 & \second{2} \\
& \textbf{PixelSR (Ours)} & 25 & 4.44 & \best{25.7157} & \best{0.8056} & \best{0.1459} & \best{56.2204} & 4.8409 & \best{75.6927} & \second{0.5612} & \second{3.9458} & \best{1} \\

\bottomrule
\end{tabular}
}
\label{quantitative_all}
\vspace{-0.6cm}
\end{table*}

\paragraph{Task-Aligned Branch Assignment.}
Motivated by this physical correspondence, OFD assigns each decoupled objective to the network branch best suited to its characteristics. Specifically, the global-focused DiT backbone is tasked with semantic prediction $\widehat{\boldsymbol{u}}_{\mathrm{sem}}$, while the local-focused Pixel Decoder targets texture prediction $\widehat{\boldsymbol{u}}_{\mathrm{tex}}$, yielding:
\begin{equation}
\widehat{\boldsymbol{u}}_{\parallel}
=
\boldsymbol{P}\widehat{\boldsymbol{u}}_{\mathrm{sem}},
\qquad
\widehat{\boldsymbol{u}}_{\perp}
=
(\boldsymbol{I}-\boldsymbol{P})
\widehat{\boldsymbol{u}}_{\mathrm{tex}}.
\label{eq:ofd_prediction}
\end{equation}
Accordingly, this task-aligned separation successfully isolates their prediction spaces, mitigating optimization interference while allowing the Pixel Decoder to leverage noise-free supervision for high-fidelity texture reconstruction.

\paragraph{Full-Rank Velocity Recovery and Training.}
Required by Conditional Flow Matching, the two branch predictions are converted into full-rank velocity. From
$\boldsymbol{x}_t=\boldsymbol{x}_0+t\boldsymbol{v}$ and Eq.~\ref{eq:ofd_decomposition}, we derive:
\begin{equation}
\widehat{\boldsymbol{v}}
=
\widehat{\boldsymbol{u}}_{\parallel}
+
\frac{1}{t}
\left[
(\boldsymbol{I}-\boldsymbol{P})\boldsymbol{x}_t
+
\widehat{\boldsymbol{u}}_{\perp}
\right].
\label{eq:branch_velocity_recovery}
\end{equation}
Finally, we optimize $\widehat{\boldsymbol{v}}$ against the full-rank target $\boldsymbol{v}$ using  standard loss in Eq.~\ref{eq:cfm_loss}. Since the two prediction errors occupy orthogonal subspaces, their inner-product cross term vanishes, allowing both branches to be jointly trained without mixing their optimization signals. At inference,  recovered velocity is directly integrated by a standard ODE solver without modifying the sampling procedure.

\section{Experiments}

\subsection{Experiment Settings}
\label{sec:exp_settings}

\paragraph{Datasets.}
We train PixelSR on LSDIR~\citep{LSDIR}, DIV2K~\citep{DIV2K}, DIV8K~\citep{DIV8K}, OutdoorSceneTraining~\citep{OST}, Flickr2K~\citep{Flickr2K}, the first 10K images from FFHQ~\citep{FFHQ}, and 50K high-quality images collected from the web. The LR--HR pairs are synthesized via the Real-ESRGAN pipeline~\citep{Real-esrgan}. 

For synthetic evaluation, we adopt the LSDIR validation split, center-cropping GT to $512\times512$ before synthesizing $128\times128$ LR inputs. 
For real-world evaluation, we utilize RealSR~\citep{RealSR}, ScreenSR~\citep{VOSR}, and our newly constructed benchmark, \textbf{UGCSR}, which contains more contemporary content and complex degradations collected from real streaming-media pipelines (details in \textbf{Appendix} \ref{sec:ugcsr_details}). 
Specifically, we directly center-crop paired $512\times512$ patches from these benchmarks for evaluation.

\begin{figure*}[t]
    \centering
    \includegraphics[width=1\linewidth]{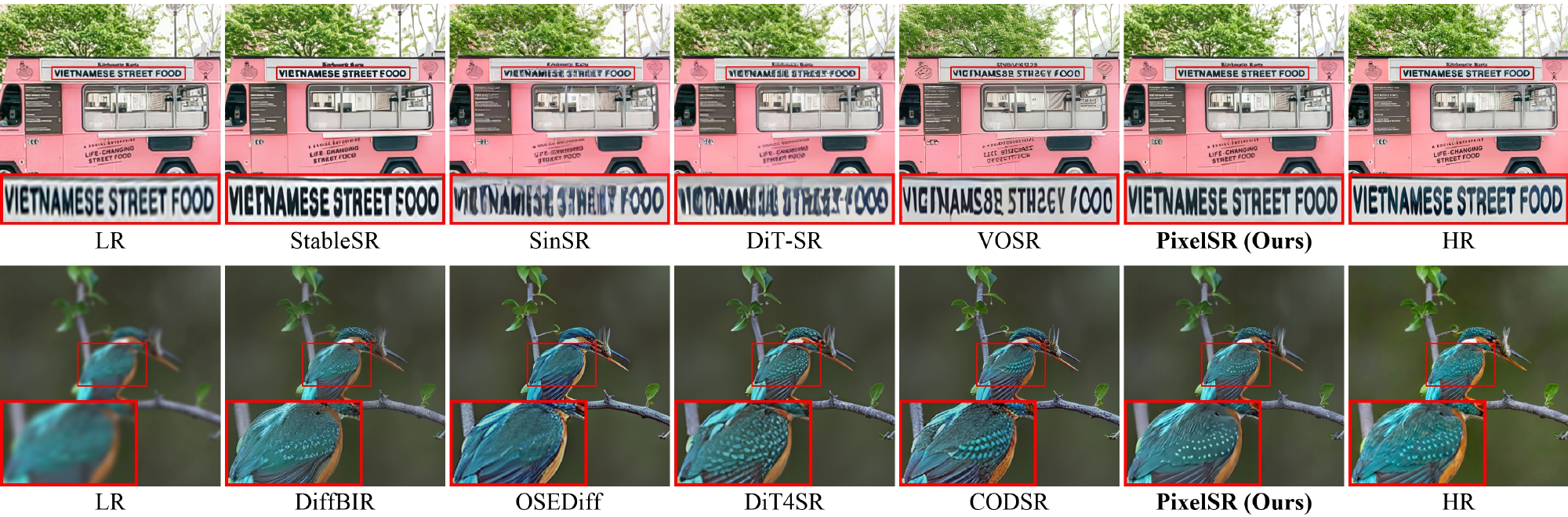}
    \vspace{-0.4cm}
    \caption{\textbf{Qualitative comparison on challenging real-world examples.} 
    PixelSR better preserves the text layout and feather structure while recovering fine-grained details consistent with the HR references, whereas several competing methods exhibit distorted characters, irregular textures, or over-smoothed results. Best viewed zoomed in.
    }
    \label{fig:compare}
    \vspace{-0.4cm}
\end{figure*}



\paragraph{Evaluation Metrics.}
We report reference-based metrics, including pixel-level PSNR and SSIM~\citep{SSIM}, perceptual LPIPS~\citep{LPIPS}, and distribution-level FID~\citep{FID}. Additionally, we employ the no-reference metrics NIQE~\citep{NIQE}, MUSIQ~\citep{MUSIQ}, MANIQA~\citep{MANIQA}, and DeQA~\citep{DeQA}. 


To facilitate holistic comparison across multiple evaluation metrics, we compute the average rank of each method over all metrics:
\begin{equation}
R_k =
\frac{1}{|\mathcal{M}|}
\sum_{m\in\mathcal{M}}
\operatorname{rank}_k(m),
\label{eq:average_rank}
\end{equation}
where $\mathcal{M}$ denotes the  metric sets. The resulting average rank, reported as ``Rank'' in Table~\ref{quantitative_all}, is obtained by sorting $R_k$ and serves as an aggregate summary of overall performance.

\paragraph{Implementation Details.} 
PixelSR is optimized end-to-end in pixel space. 
The patch-level backbone consists of $N=28$ DiT blocks with hidden dimension $D_{\mathrm{DiT}}=1024$, while the pixel-level decoder comprises $M=4$ PiT blocks with hidden dimension $D_{\mathrm{PiT}}=32$. 
We use a frozen pre-trained DINOv2-B/14~\citep{Dinov2} as the visual semantic encoder $E_{\mathrm{vis}}$. 
Following the rank analysis in AsymFlow~\citep{AsymFlow}, we fix $r=16$ across all experiments. This choice also matches the $2\times2\times4=16$ latent coefficients corresponding to each $16\times16$ RGB patch under an $f8c4$ VAE, facilitating dimensionally aligned Procrustes fitting.
Please see \textbf{Appendix} \ref{sec:more_implementation} for further implementation details.

\subsection{Comparison with State-of-the-Art}

\paragraph{Compared Methods.}
We compare PixelSR with recent diffusion-based Real-SR methods, including DiffBIR~\citep{DiffBIR}, StableSR~\citep{StableSR}, OSEDiff~\citep{OSEDiff}, SinSR~\citep{SinSR}, DiT4SR~\citep{DiT4SR}, DiT-SR~\citep{DiTSR}, CODSR~\citep{CODSR}, and VOSR~\citep{VOSR}. All competing methods are evaluated using their officially released checkpoints and default inference settings. Notably, these methods operate in compressed latent spaces, whereas PixelSR performs restoration directly in RGB space.

\paragraph{Quantitative Comparisons.}
Table~\ref{quantitative_all} reports the quantitative results on four benchmarks. Existing methods exhibit different preferences along the fidelity--realism spectrum. For example, SinSR and DiT-SR achieve competitive PSNR and SSIM on several benchmarks but are less favorable on perceptual-quality metrics. Conversely, DiffBIR, CODSR, and VOSR attain strong no-reference scores in certain cases, often accompanied by degraded reference-based fidelity. 
PixelSR achieves the best overall rank across the four benchmarks while maintaining competitive performance on both metric groups. In particular, it obtains substantial fidelity improvements on the challenging UGCSR benchmark without sacrificing perceptual quality. Together with the trade-off analysis in Figure~\ref{fig:teaser}, these results demonstrate that our PixelSR effectively alleviates the fidelity--realism \textit{seesaw effect}.

\paragraph{Qualitative Comparisons.}
Figure~\ref{fig:compare} presents qualitative comparisons on challenging real-world cases. In the first example, competing methods distort character strokes or hallucinate incorrect layouts, whereas PixelSR preserves text structures consistent with the HR reference. In the second example, competing methods tend to produce over-smoothed or irregular feather patterns. PixelSR reconstructs finer and more natural textures without altering the bird's appearance. 
These visual results further corroborate the quantitative evidence that our single-stage, end-to-end  PixelSR better reconciles structural fidelity and perceptual realism.

\paragraph{Efficiency Comparisons.}
We evaluate all methods on a single NVIDIA H800 GPU using their official sampling settings. As reported in Table~\ref{quantitative_all}, PixelSR requires approximately 4.5 seconds per image, with runtime comparable to several multi-step latent diffusion baselines. This work focuses on restoration quality and the fidelity–perception trade-off, while integrating one-step distillation for further acceleration is left for future exploration.


\subsection{Ablation Study}
\label{sec:ablation}

\begin{figure}[t]
    \centering
    \includegraphics[width=\linewidth]{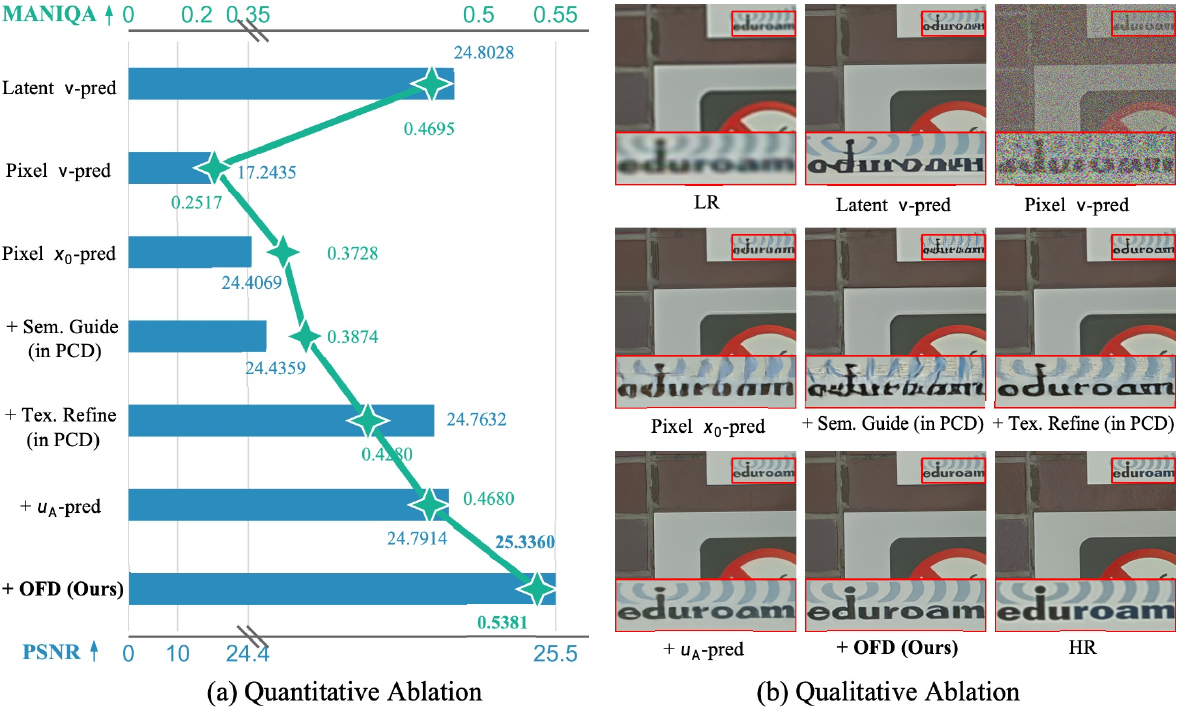}
    \caption{\textbf{Ablation of PixelSR components on RealSR.}
    The bars and green curve report PSNR and MANIQA, respectively.
    Pixel-space $\boldsymbol{x}_0$-prediction provides a substantially
    stronger baseline than pixel-space $\boldsymbol{v}$-prediction,
    while the successive PCD and OFD designs further improve fidelity
    and perceptual quality.}
    \label{fig:core}
\end{figure}

\paragraph{Effectiveness of Core Components.}
Figure~\ref{fig:core} presents a progressive ablation from a standard latent DiT baseline to the complete PixelSR under a shared training protocol. 
Directly transferring $\boldsymbol{v}$-prediction from latent to pixel space while retaining the DiT configuration causes a severe performance drop, highlighting the difficulty of predicting full-dimensional noise in RGB space. 
Replacing it with $\boldsymbol{x}_0$-prediction yields a substantially stronger pixel-space baseline. 
On this basis, semantic guidance improves perceptual quality while preserving fidelity, and the PiT-based texture refinement further enhances local details, producing clear gains in both PSNR and MANIQA. 
Restricting noise prediction to a low-rank subspace through the asymmetric target $\boldsymbol{u}_{\mathrm{A}}$ provides additional improvements. 
Finally, OFD assigns the complementary objectives to specialized branches, achieving the best PSNR and MANIQA among all variants. 
The qualitative progression further illustrates increasingly faithful text structures and sharper local details.

\begin{figure}[t]
    \centering
    \includegraphics[width=\linewidth]{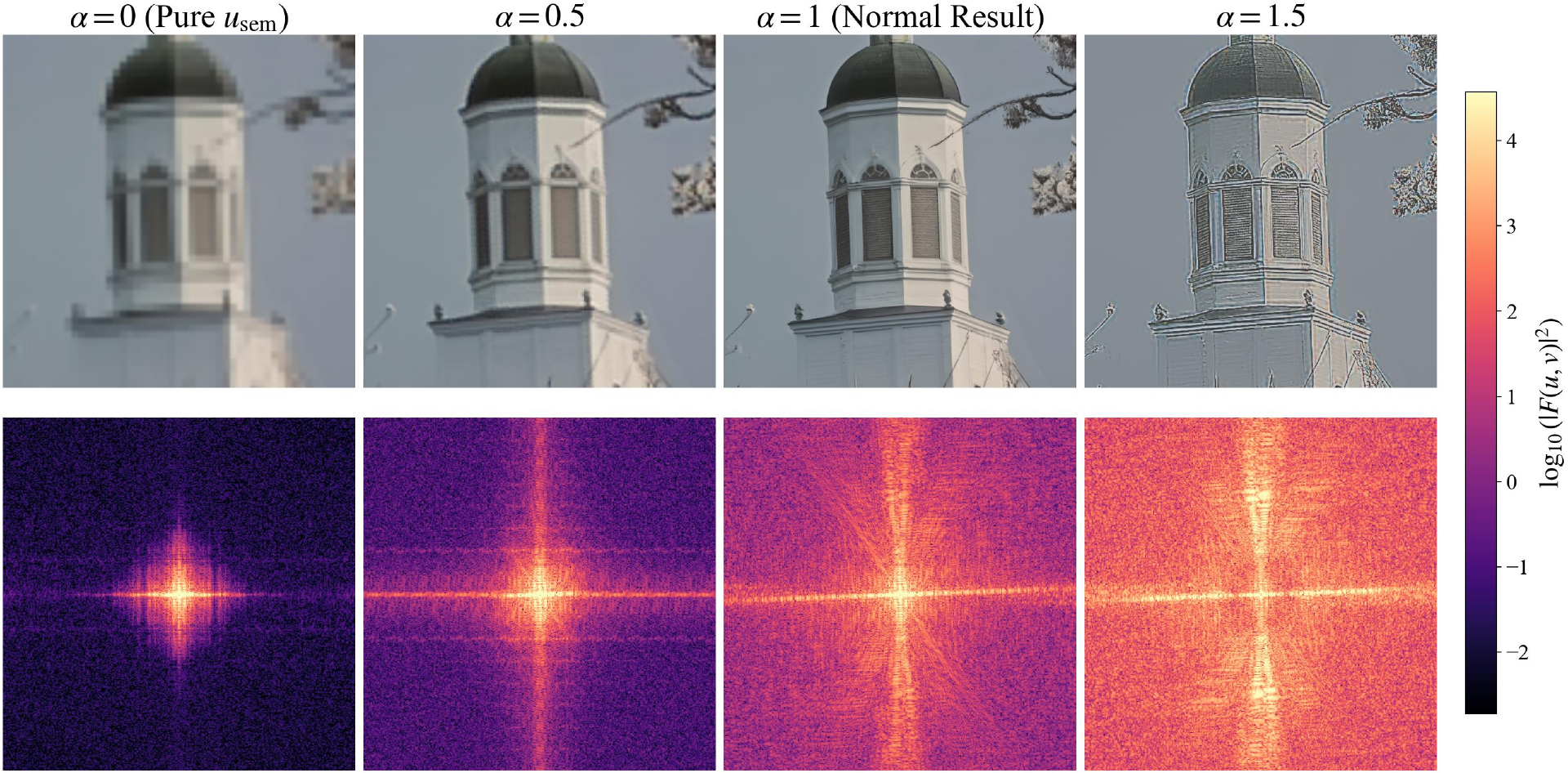}
    \caption{\textbf{Frequency analysis of OFD branch specialization.}
    Restoration results (top) and corresponding power spectra (bottom)
    under increasing orthogonal scale $\alpha$. Larger $\alpha$
    progressively strengthens fine textures and mid-to-high-frequency
    responses, while $\alpha=1$ denotes the default OFD output.
    All spectra share the same color scale.}
    \label{fig:frequency_analysis}
\end{figure}


\begin{wraptable}{r}{0.47\textwidth}
\vspace{-0.4cm}
\centering
\caption{\textbf{Ablation of pixel-space flow formulations on LSDIR.}
All variants use the same PCD network and parameter count.
OFD improves over monolithic $\boldsymbol{u}_{\mathrm A}$-prediction
through explicit factorization and branch assignment.}
\vspace{-0.2cm}
\setlength{\tabcolsep}{2pt}
\renewcommand{\arraystretch}{1.3}
\resizebox{\linewidth}{!}{
\begin{tabular}{ccccccc}
\toprule
\multirow{2}{*}{Variant}
& \multicolumn{3}{c}{Reference-based Metrics}
& \multicolumn{3}{c}{No-Reference Metrics} \\
\cmidrule(lr){2-4} \cmidrule(lr){5-7}
& PSNR $\uparrow$ & SSIM $\uparrow$ & LPIPS $\downarrow$
& MUSIQ $\uparrow$ & MANIQA $\uparrow$ & DeQA $\uparrow$ \\
\midrule
$\boldsymbol{v}$-prediction
& 15.0173 & 0.0995 & 1.2483
& 26.5085 & 0.2176 & 1.5582 \\

$\boldsymbol{x}_0$-prediction
& 19.5442 & 0.4822 & 0.2543
& 72.3934 & 0.4895 & 3.6563 \\

$\boldsymbol{u}_{\mathrm A}$-prediction
& \second{20.3088} & \second{0.5101} & \second{0.2251}
& \second{72.4436} & \second{0.5495} & \second{3.9879} \\

\textbf{OFD (Ours)}
& \best{20.6907} & \best{0.5270} & \best{0.2203}
& \best{73.1895} & \best{0.5661} & \best{4.0936} \\
\bottomrule
\end{tabular}
}
\label{tab:lsdir_flow_ablation}
\vspace{-0.15cm}
\end{wraptable}
\paragraph{Analysis of Orthogonal Flow Decoupling.}
To isolate the contribution of OFD, we compare four flow formulations using the same PCD network, training settings, and parameter count. For the monolithic $\boldsymbol{v}$-, $\boldsymbol{x}_0$-, and $\boldsymbol{u}_{\mathrm A}$-prediction variants, the final PiT output directly predicts the complete target. OFD introduces no additional modules, but instead assigns the projected and orthogonal objectives to the DiT and PiT outputs, respectively.
As shown in Table~\ref{tab:lsdir_flow_ablation}, $\boldsymbol{x}_0$-prediction substantially improves over standard $\boldsymbol{v}$-prediction, while monolithic $\boldsymbol{u}_{\mathrm A}$-prediction provides further gains by restricting noise prediction to a low-rank subspace. OFD consistently performs best across all metrics. Since it differs from monolithic $\boldsymbol{u}_{\mathrm A}$-prediction only in explicit factorization and branch assignment, the gains isolate the effectiveness of OFD rather than additional model capacity.


We further probe the functional specialization of the two OFD branches by scaling the orthogonal texture contribution at validation time:
\begin{equation}
\widehat{\boldsymbol{v}}_{\alpha}
=
\widehat{\boldsymbol{u}}_{\parallel}
+
\frac{\alpha}{t}
\left[
(\boldsymbol{I}-\boldsymbol{P})\boldsymbol{x}_t
+
\widehat{\boldsymbol{u}}_{\perp}
\right].
\end{equation}
Here, $\alpha=0$ retains only the projected semantic component, whereas $\alpha=1$ recovers the standard OFD formulation. As shown in Figure~\ref{fig:frequency_analysis}, increasing $\alpha$ preserves the global layout while progressively sharpening local structures and strengthening mid- and high-frequency responses; an excessive scale ($\alpha=1.5$) produces over-sharpening artifacts. This controlled intervention provides frequency-domain evidence that the projected branch primarily maintains global structure, whereas the orthogonal branch supplies fine-grained textures. The scale analysis is used only for interpretation, and $\alpha=1$ is fixed for all reported results.

\section{Conclusion}
In this paper, we rethink the fidelity--realism trade-off in Real-SR by challenging the prevailing latent diffusion paradigm. To bypass lossy VAE compression and cascading errors, we present PixelSR, a single-stage framework operating directly in RGB space. Crucially, to resolve task entanglement in pixel-space flow matching, we introduce Orthogonal Flow Decoupling, which mathematically decomposes asymmetric velocity prediction into semantic generation and deterministic texture recovery assigned to specialized sub-networks. Extensive experiments demonstrate that our PixelSR substantially alleviates the fidelity--realism \textit{seesaw effect} and establishes a more favorable Pareto frontier. Beyond super-resolution, our work validates the viability of pixel diffusion without VAE encoding and decoding, providing a principled baseline for harmonizing generative priors with strict structural fidelity.

\clearpage

\subsection*{AI use statement}
In this work, we used generative AI tools to polish the content. We did not use generative AI tools
for data generation or concept design. We have reviewed all work completed with the assistance
of AI. We take responsibility for the final content of this work, including text, claims or artifacts produced with the aid of generative AI.

\subsection*{Reproducibility statement}
The code will be released after the paper is accepted.
\bibliography{iclr2027_conference}
\bibliographystyle{iclr2027_conference}

\clearpage
\appendix
\section{Appendix}
\subsection{More Implementation Details}
\label{sec:more_implementation}

Table~\ref{tab:implementation_details} summarizes the detailed  architectural specifications and training settings of our PixelSR. Unless otherwise specified, all experiments in both the main paper and the appendix adopt this default configuration.
\begin{table}[h!]
\vspace{-0.1cm}
\centering
\caption{Implementation details of PixelSR.}
\small
\renewcommand{\arraystretch}{1.08}
\setlength{\tabcolsep}{3pt}
\begin{tabular}{lc}
\toprule
\textbf{Hyperparameter} & \textbf{PixelSR} \\
\midrule

\multicolumn{2}{l}{\textit{Architecture}} \\
Restoration resolution
& $512^{2}\times3$ \\
Conditional input dimension
& $512^{2}\times6$ \\
Patch-level depth $N$
& $28$ \\
Pixel-level depth $M$
& $4$ \\
Hidden size $D_{\mathrm{DiT}}$
& $1024$ \\
Patch-level attention heads
& $16$ \\
Pixel hidden size $D_{\mathrm{PiT}}$
& $32$ \\
Pixel-level attention heads
& $8$ \\
Patch size $p$
& $16$ \\
Pixel tokens per patch
& $256$ \\
MLP ratio
& $4$ \\
\#Params (M)
& $773.4$ \\

\midrule
\multicolumn{2}{l}{\textit{Visual Semantic Guidance}} \\
Semantic encoder
& DINOv2-B/14~\citep{Dinov2} \\
Semantic encoder input
& $448^{2}\times3$ \\
Semantic injection
& Cross-attention \\

\midrule
\multicolumn{2}{l}{\textit{Orthogonal Flow Decoupling}} \\
Projection rank $r$
& $16$ \\
Projection basis
& Procrustes alignment \\
Minimum noise scale  & $0.05$ \\

\midrule
\multicolumn{2}{l}{\textit{Optimization}} \\
Training iterations
& $500$K \\
Batch size
& $8$ \\
Optimizer
& Adam \\
Adam coefficients $(\beta_1,\beta_2)$
& $(0.9,\,0.95)$ \\
Learning rate
& $5\times10^{-5}$ \\
Weight decay
& $1\times10^{-2}$ \\
Learning-rate schedule
& Constant \\
Gradient clipping
& $1.0$ \\
Mixed precision
& BF16 \\
EMA decay
& $0.9999$ \\
Timestep distribution
& $\mathcal{U}(0,1)$ \\
Random seed
& $42$ \\

\midrule
\multicolumn{2}{l}{\textit{Inference}} \\
Sampler & Euler-based ODE solver \\
Sampling steps
& $25$ \\
Restoration space
& RGB pixel space \\

\bottomrule
\end{tabular}

\vspace{-0.5cm}
\label{tab:implementation_details}
\end{table}

\subsection{UGCSR Benchmark Details}
\label{sec:ugcsr_details}

\paragraph{Benchmark Construction and Content Diversity.}
UGCSR is a paired benchmark for evaluating real-world generative SR on contemporary user-generated content (UGC). It contains $230$ spatially aligned LR--HR pairs obtained through real streaming-media processing pipelines. The original high-quality images are retained as HR references, while their platform-processed counterparts serve as realistic LR inputs. All pairs are cropped into $512\times512$ patches and used exclusively for evaluation.

\begin{figure*}[t]
    \centering
    \includegraphics[width=1\linewidth]{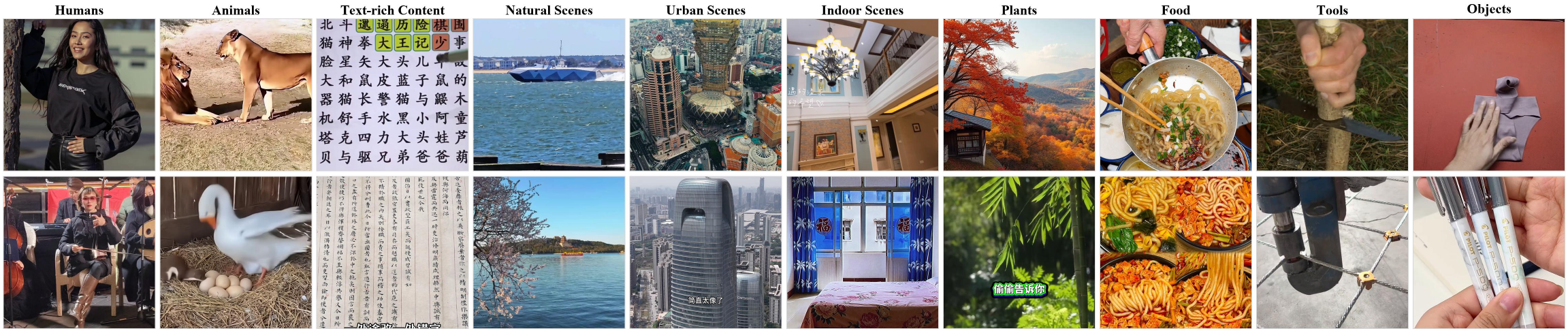}
\caption{\textbf{Representative image matrix of UGCSR benchmark.}
The benchmark covers ten semantic categories: humans, animals,
text-rich images, natural scenes, urban scenes, indoor scenes, plants,
food, tools, and everyday objects. Two representative degraded images
are shown for each category.}
    \label{fig:ugcsr_content}
\end{figure*}

UGCSR covers nine semantic categories, including humans, animals, text-rich images, natural scenes, urban scenes, indoor scenes, food, tools, and everyday objects. As illustrated in Figure~\ref{fig:ugcsr_content}, the samples exhibit diverse object scales, layouts, illumination conditions, and texture complexities.
Unlike camera-captured benchmarks, UGCSR captures compound, content-dependent degradations introduced by practical media
distribution, such as resolution adaptation, compression, transcoding, blur, ringing, aliasing, noise, and color variations.

\begin{figure*}[t]
    \centering
    \includegraphics[width=1\linewidth]{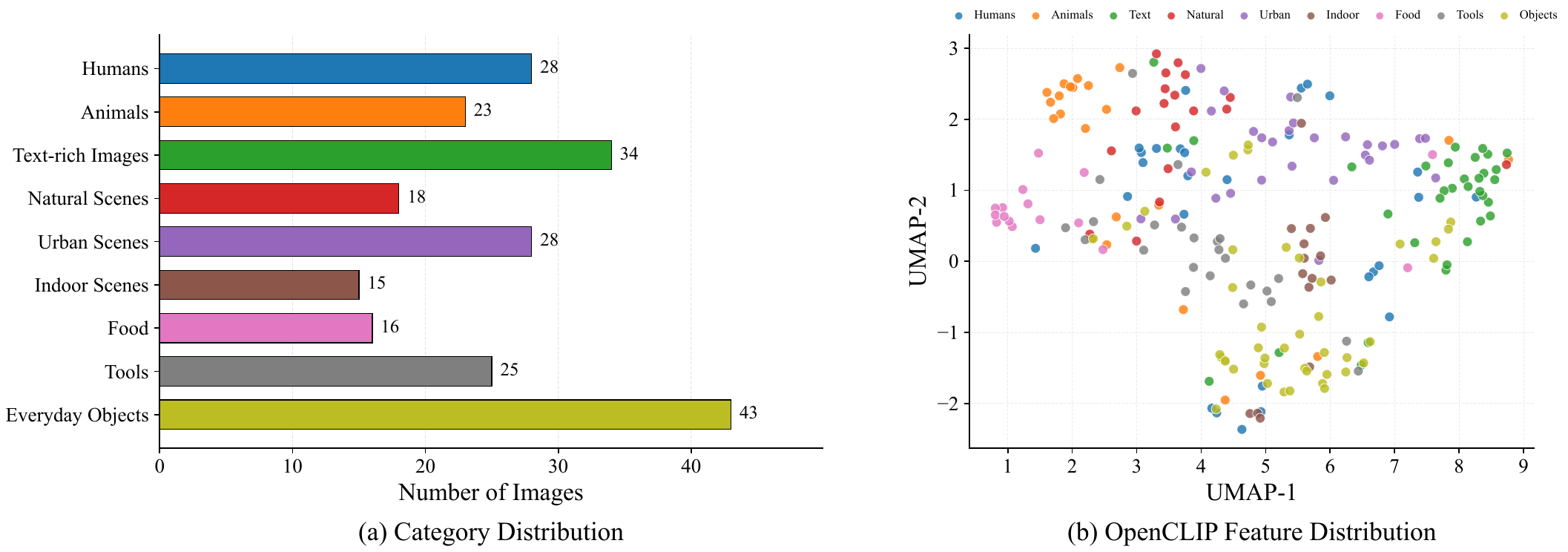}
\caption{\textbf{Objective semantic analysis of UGCSR.}
(a) Distribution of OpenCLIP zero-shot pseudo-labels across nine
semantic categories. (b) UMAP visualization of the OpenCLIP image
features, where each point represents one HR image and colors indicate
the corresponding Top-1 predicted categories.}
    \label{fig:ugcsr_semantic_diversity}
    \vspace{-0.3CM}
\end{figure*}

\paragraph{Objective Semantic Analysis.}
To objectively assess the content diversity of UGCSR, we employ OpenCLIP~\citep{clip,OpenCLIP} to perform zero-shot classification on all $230$ images using ten predefined semantic categories. Each image is assigned to the category whose text prototype has the highest cosine similarity with its normalized image feature. As shown in Figure~\ref{fig:ugcsr_semantic_diversity}(a), the resulting distribution covers humans, animals, text-rich images, natural scenes, urban scenes, indoor scenes, plants, food, tools, and everyday objects, demonstrating the broad content coverage of UGCSR.

We further project the OpenCLIP features into two dimensions using UMAP~\citep{Umap}, as shown in Figure~\ref{fig:ugcsr_semantic_diversity}(b). The locally clustered yet  overlapping distribution further reflects the semantic diversity and non-exclusive nature of real-world UGC content.

\paragraph{Quality Assessment Analysis.}
Table~\ref{tab:dataset_quality} reports the mean and standard deviation of four no-reference image quality metrics for both LR and HR images.
UGCSR achieves the best HR scores on MUSIQ, MANIQA, and DeQA, together with a competitive NIQE score. Its relatively small standard deviations further indicate consistently high-quality references, supporting reliable paired evaluation.

Meanwhile, clear LR--HR quality gaps are observed across all four metrics, confirming the presence of perceptible degradations introduced
by real media-processing pipelines. Compared with RealSR and ScreenSR, UGCSR retains higher-quality LR inputs, suggesting that it emphasizes
realistic and relatively subtle degradations rather than severely corrupted observations.

\begin{table*}[t]
\centering
\footnotesize
\caption{No-reference quality comparison of LR and HR images across
real-world paired SR benchmarks. Results are reported as mean $\pm$
standard deviation. The best results among LR and HR images are
highlighted in \textbf{bold} and \underline{underlined}, respectively.}
\setlength{\tabcolsep}{5pt}
\renewcommand{\arraystretch}{1.15}
\resizebox{\textwidth}{!}{
\begin{tabular}{lcccccc}
\toprule
\multirow{2}{*}{Metric}
& \multicolumn{2}{c}{RealSR}
& \multicolumn{2}{c}{ScreenSR}
& \multicolumn{2}{c}{UGCSR} \\
\cmidrule(lr){2-3}
\cmidrule(lr){4-5}
\cmidrule(lr){6-7}
& LR & HR & LR & HR & LR & HR \\
\midrule

MUSIQ $\uparrow$
& 29.597 $\pm$ 6.335
& 57.465 $\pm$ 9.239
& 41.220 $\pm$ 12.321
& 72.150 $\pm$ 4.672
& \textbf{69.753 $\pm$ 9.747}
& \underline{75.175 $\pm$ 3.196} \\

MANIQA $\uparrow$
& 0.278 $\pm$ 0.084
& 0.329 $\pm$ 0.093
& 0.212 $\pm$ 0.081
& 0.565 $\pm$ 0.061
& \textbf{0.464 $\pm$ 0.112}
& \underline{0.588 $\pm$ 0.056} \\

DeQA $\uparrow$
& 1.721 $\pm$ 0.186
& 3.050 $\pm$ 0.418
& 2.341 $\pm$ 0.403
& 4.121 $\pm$ 0.303
& \textbf{3.559 $\pm$ 0.537}
& \underline{4.183 $\pm$ 0.215} \\

NIQE $\downarrow$
& 18.215 $\pm$ 28.351
& 6.117 $\pm$ 1.444
& 6.968 $\pm$ 1.248
& \underline{3.772 $\pm$ 1.952}
& \textbf{4.797 $\pm$ 1.841}
& 3.884 $\pm$ 1.428 \\
\bottomrule
\end{tabular}
}
\vspace{-0.4cm}
\label{tab:dataset_quality}

\end{table*}

\subsection{Low-Rank Projector Construction}
\label{sec:projector_construction}

OFD relies on a rank-$r$ orthogonal projector
$\boldsymbol{P}\in\mathbb{R}^{D\times D}$ to partition each
$D$-dimensional pixel patch into two complementary subspaces. We
parameterize $\boldsymbol{P}$ using an orthonormal basis
$\boldsymbol{A}\in\mathbb{R}^{D\times r}$ satisfying:
\begin{equation}
    \boldsymbol{A}^{\mathsf{T}}\boldsymbol{A}
    =
    \boldsymbol{I}_{r}.
\end{equation}
The corresponding projector is defined as:
\begin{equation}
    \boldsymbol{P}
    :=
    \boldsymbol{A}\boldsymbol{A}^{\mathsf{T}}.
\end{equation}
It satisfies
$\boldsymbol{P}^{\mathsf{T}}=\boldsymbol{P}$,
$\boldsymbol{P}^{2}=\boldsymbol{P}$, and
$\operatorname{rank}(\boldsymbol{P})=r$. Moreover, the orthonormality
constraint preserves the Gaussian distribution of projected
pixel-space noise. Specifically, if
$\boldsymbol{\epsilon}\sim\mathcal{N}(\boldsymbol{0},
\boldsymbol{I}_{D})$, then:
\begin{equation}
    \boldsymbol{A}^{\mathsf{T}}\boldsymbol{\epsilon}
    \sim
    \mathcal{N}(\boldsymbol{0},\boldsymbol{I}_{r}).
\end{equation}
We investigate three strategies for constructing the projector.

\paragraph{Learnable Projector $\boldsymbol{P}$.}
The first strategy constructs $\boldsymbol{P}$ from a learnable
orthonormal basis $\boldsymbol{A}$, which is jointly optimized with
PixelSR under the following constraint:
\begin{equation}
    \boldsymbol{A}^{\mathsf{T}}\boldsymbol{A}
    =
    \boldsymbol{I}_{r}.
\end{equation}
The resulting projector is:
\begin{equation}
    \boldsymbol{P}_{\mathrm{Learn}}
    :=
    \boldsymbol{A}\boldsymbol{A}^{\mathsf{T}}.
\end{equation}
This formulation allows the projected subspace to adapt directly to the restoration objective. However, the projector continuously changes during training and does not explicitly incorporate prior knowledge about the representation structure of natural images.

\paragraph{SVD-Based Projector $\boldsymbol{P}$.}
The second strategy constructs a data-dependent projector from the principal directions of pixel patches. Specifically, we collect $K$ normalized and vectorized RGB patches and stack them as:
\begin{equation}
    \boldsymbol{X}
    =
    \left[
        \boldsymbol{x}_{1},
        \ldots,
        \boldsymbol{x}_{K}
    \right]^{\mathsf{T}}
    \in
    \mathbb{R}^{K\times D}.
\end{equation}
For a patch size of $p$, the pixel dimension is:
\begin{equation}
    D=3p^{2}.
\end{equation}
Applying singular value decomposition to the centered pixel matrix
gives:
\begin{equation}
    \boldsymbol{X}
    =
    \boldsymbol{U}_{x}
    \boldsymbol{\Sigma}_{x}
    \boldsymbol{V}_{x}^{\mathsf{T}}.
\end{equation}
The top-$r$ right singular vectors define the principal pixel
subspace:
\begin{equation}
    \boldsymbol{A}_{\mathrm{SVD}}
    :=
    \boldsymbol{V}_{x}[:,1:r].
\end{equation}
The corresponding SVD-based projector is:
\begin{equation}
    \boldsymbol{P}_{\mathrm{SVD}}
    :=
    \boldsymbol{A}_{\mathrm{SVD}}
    \boldsymbol{A}_{\mathrm{SVD}}^{\mathsf{T}}.
\end{equation}
This construction retains the dominant variations of natural image patches without introducing an additional learned representation. Nevertheless, directions with large pixel variance are not necessarily aligned with the compact semantic organization required by OFD.

\paragraph{Procrustes-Aligned Projector $\boldsymbol{P}$.}
To construct a semantically meaningful projector, we align each $p\times p$ RGB patch $\boldsymbol{x}_{i}\in\mathbb{R}^{D}$ with its corresponding VAE latent block $\boldsymbol{z}_{i}\in\mathbb{R}^{r}$.  Specifically, we use the $f8c4$ VAE from Stable Diffusion 2.1 (SD2.1) by default. With $p=16$, the dimensions are:
\begin{equation}
    D=3\times16^{2}=768,
    \qquad
    r=2\times2\times4=16.
\end{equation}
We stack the paired pixel and latent vectors as:
\begin{equation}
\begin{aligned}
    \boldsymbol{X}
    &=
    \left[
        \boldsymbol{x}_{1},
        \ldots,
        \boldsymbol{x}_{K}
    \right]^{\mathsf{T}}
    \in
    \mathbb{R}^{K\times D}, \\
    \boldsymbol{Z}
    &=
    \left[
        \boldsymbol{z}_{1},
        \ldots,
        \boldsymbol{z}_{K}
    \right]^{\mathsf{T}}
    \in
    \mathbb{R}^{K\times r}.
\end{aligned}
\end{equation}
Then, we seek a semi-orthogonal mapping that lifts the compact latent
representation into the pixel-patch space:
\begin{equation}
    \boldsymbol{A}^{*}
    :=
    \arg\min_{\boldsymbol{A}^{\mathsf{T}}\boldsymbol{A}
    =\boldsymbol{I}_{r}}
    \left\|
        \boldsymbol{X}
        -
        \boldsymbol{Z}\boldsymbol{A}^{\mathsf{T}}
    \right\|_{\mathrm{F}}^{2}.
    \label{eq:procrustes_objective}
\end{equation}
The objective in Eq.~\ref{eq:procrustes_objective} can be rewritten as:
\begin{equation}
    \left\|
        \boldsymbol{X}
        -
        \boldsymbol{Z}\boldsymbol{A}^{\mathsf{T}}
    \right\|_{\mathrm{F}}^{2}
    =
    \left\|\boldsymbol{X}\right\|_{\mathrm{F}}^{2}
    +
    \left\|\boldsymbol{Z}\right\|_{\mathrm{F}}^{2} 
    -
    2\operatorname{Tr}
    \left(
        \boldsymbol{A}^{\mathsf{T}}
        \boldsymbol{X}^{\mathsf{T}}
        \boldsymbol{Z}
    \right).
\end{equation}
Since the first two terms are independent of $\boldsymbol{A}$, the
problem is equivalent to:
\begin{equation}
    \boldsymbol{A}^{*}
    :=
    \arg\max_{\boldsymbol{A}^{\mathsf{T}}\boldsymbol{A}
    =\boldsymbol{I}_{r}}
    \operatorname{Tr}
    \left(
        \boldsymbol{A}^{\mathsf{T}}
        \boldsymbol{X}^{\mathsf{T}}
        \boldsymbol{Z}
    \right).
\end{equation}
The thin SVD of the cross-covariance matrix is defined as:
\begin{equation}
    \boldsymbol{X}^{\mathsf{T}}\boldsymbol{Z}
    =
    \boldsymbol{U}
    \boldsymbol{\Sigma}
    \boldsymbol{V}^{\mathsf{T}},
\end{equation}
where
$\boldsymbol{U}\in\mathbb{R}^{D\times r}$ and
$\boldsymbol{V}\in\mathbb{R}^{r\times r}$. The closed-form
Procrustes solution is:
\begin{equation}
    \boldsymbol{A}^{*}
    :=
    \boldsymbol{U}\boldsymbol{V}^{\mathsf{T}}.
\end{equation}
Therefore, the Procrustes-aligned projector used by OFD is:
\begin{equation}
    \boldsymbol{P}_{\mathrm{Proc}}
    :=
    \boldsymbol{A}^{*}
    \boldsymbol{A}^{*\mathsf{T}}
    =
    \boldsymbol{U}
    \boldsymbol{U}^{\mathsf{T}}.
\label{eq:procrustes_projector}
\end{equation}

Unlike the SVD-based projector, the Procrustes-aligned projector captures pixel-space directions correlated with compact VAE representations rather than dominant pixel variance. It performs best in our ablation study and is therefore adopted in OFD. Crucially, the projector $\boldsymbol{P}$ is computed offline and fixed, while the VAE is absent from PixelSR training and inference.

\paragraph{Ablation on Projection Strategies.}
Since the functional roles of the induced subspaces depend on the projection basis, we compare three rank-matched constructions of $\boldsymbol{P}$ under the same PCD architecture and training settings: a learnable orthonormal basis optimized with the restoration objective, an SVD basis derived from dominant pixel-patch variations, and a Procrustes basis aligned with corresponding pre-trained VAE tokens. 
As shown in Table~\ref{tab:screensr_p_ablation}, the Procrustes projector achieves the best results across all metrics, suggesting that the compact organization of VAE representations provides a useful prior for defining the low-rank subspace.

Furthermore, to examine whether restoration performance depends on the source VAE's reconstruction quality, we compare alignment using SD2.1, SD3, and FLUX.2 VAEs, adapting the projection-layer input dimensions to their respective latent representations.
As shown in Table~\ref{tab:screensr_p_ablation}, the three variants achieve broadly comparable SR performance, with no source VAE consistently outperforming the others across all metrics.
These results suggest that the benefit of Procrustes alignment is not specific to a particular VAE.
This is consistent with the VAE's limited role as an offline reference for constructing $\boldsymbol{P}$: PixelSR models both $\boldsymbol{P}$ and $\boldsymbol{I}-\boldsymbol{P}$ components directly in RGB space, rather than passing images through a VAE encoding--decoding bottleneck during restoration training or inference.

\subsection{Analysis of Orthogonal Flow Decoupling}
\begin{figure}[t]
    \centering
    \includegraphics[width=0.8\linewidth]{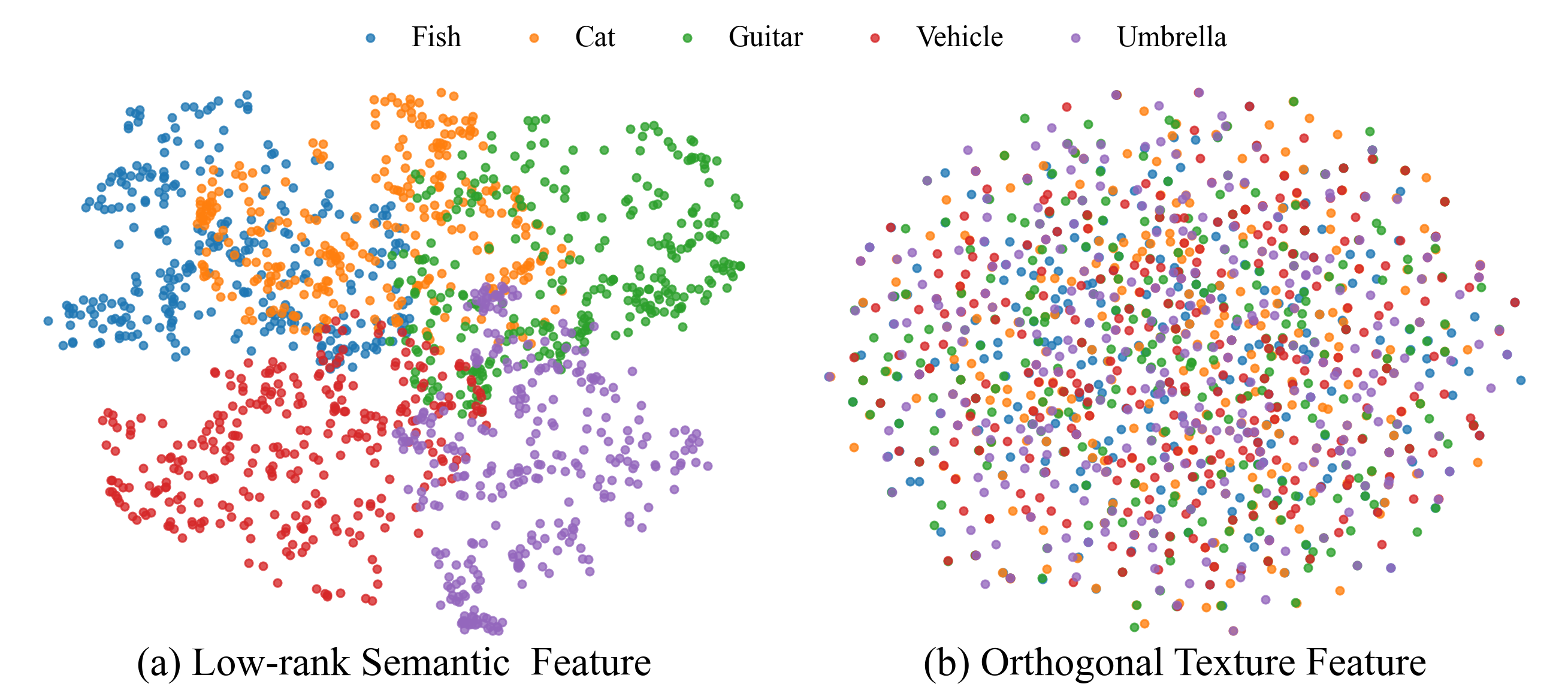}
\caption{\textbf{t-SNE visualization of semantic and texture features.}
We sample 300 images from each of five ImageNet categories and visualize features from the two PixelSR branches.
Semantic features form compact category-aware clusters, while texture features are more dispersed due to diverse local details, supporting the effectiveness of our OFD.}
    \label{fig:tsne}
    \vspace{-0.5cm}
\end{figure}

To further examine the branch specialization induced by OFD, we visualize their feature distributions using t-SNE on 300 images from each of five ImageNet~\citep{imagenet} categories, as shown in Figure~\ref{fig:tsne}.
The semantic-branch features form compact, category-aware clusters, indicating their sensitivity to high-level content.
In contrast, the texture-branch features exhibit a more dispersed distribution, consistent with the diverse local variations modeled by this branch.
Together with the component visualizations in  Figure \ref{fig:visualization} and Figure \ref{fig:frequency_analysis}, these results further provide complementary evidence for the semantic--texture interpretation underlying OFD.


\begin{table}[t]
\centering
\caption{\textbf{Ablation of projector construction on ScreenSR.}
The first three rows compare projection strategies, while the
additional rows examine different source VAEs for Procrustes alignment.
All variants use the same training settings.
\colorbox{red!40}{Red} and \colorbox{orange!40}{orange} indicate the best
and second-best performance, respectively.}
\renewcommand{\arraystretch}{1.08}
\resizebox{\linewidth}{!}{
\begin{tabular}{lcccccc}
\toprule
\multirow{2}{*}{Variant}
& \multicolumn{3}{c}{Reference-based Metrics}
& \multicolumn{3}{c}{No-Reference Metrics} \\
\cmidrule(lr){2-4} \cmidrule(lr){5-7}
& PSNR $\uparrow$ & SSIM $\uparrow$ & LPIPS $\downarrow$
& MUSIQ $\uparrow$ & MANIQA $\uparrow$ & DeQA $\uparrow$ \\
\midrule
Learnable $\boldsymbol{P}$
& 23.4996 & 0.7114 & 0.2222
& 71.8488 & 0.4687 & 3.7217 \\

SVD-based $\boldsymbol{P}$
& 23.5053 & 0.7010 & 0.2224
& 71.9188 & 0.4645 & 3.7063 \\

Procrustes $\boldsymbol{P}$ (SD2.1)
& \second{23.5640} & 0.7156 & \best{0.2082}
& \second{72.8613} & \best{0.4824} & \second{3.8294} \\
\midrule
Procrustes $\boldsymbol{P}$ (SD3)
& 23.5583 & \second{0.7160} & 0.2132
& \best{72.8709} & \second{0.4816} & 3.8146 \\

Procrustes $\boldsymbol{P}$ (FLUX.2)
& \best{23.5661} & \best{0.7168} & \second{0.2099}
& 72.8436 & 0.4809 & \best{3.8318} \\
\bottomrule
\end{tabular}
}
\label{tab:screensr_p_ablation}
\end{table}

\subsection{Additional Ablation Results}
\label{sec:additional_ablation}
We investigate the effect of different pre-trained semantic encoders by replacing DINOv2 with CLIP \citep{clip} and SigLIP \citep{siglip} while keeping all other settings unchanged. 
As reported in Table~\ref{tab:semantic_encoder_ablation}, DINOv2 achieves the best overall performance, indicating that its dense visual representations provide more effective semantic guidance for PixelSR. Therefore, we adopt DINOv2 as the default semantic encoder.

\begin{table}[h]

\centering
\footnotesize
\caption{\textbf{Ablation on different semantic encoders on RealSR.} All variants use the same PCD framework and settings. \colorbox{red!40}{Red} indicates the best performance.}
\vspace{-0.2cm}
\renewcommand{\arraystretch}{1.08}
\resizebox{\linewidth}{!}{
\begin{tabular}{lcccccc}
\toprule
Encoder
& PSNR $\uparrow$
& SSIM $\uparrow$
& LPIPS $\downarrow$
& MUSIQ $\uparrow$
& MANIQA $\uparrow$
& DeQA $\uparrow$ \\
\midrule
CLIP   & 25.2911 & 0.7161 & 0.2924 & 68.4463 & 0.5056 & 3.2331 \\
SigLIP & 25.1778 & 0.7099 & 0.2967 & 68.7191 & 0.5168 & 3.2396 \\ 
DINOv2 & \best{25.3360} & \best{0.7166} & \best{0.2922}
& \best{69.5941} & \best{0.5381} & \best{3.6116} \\ 
\bottomrule
\end{tabular}
}
\label{tab:semantic_encoder_ablation}
\label{-0.8cm}
\end{table}

\subsection{ Additional Qualitative Results}
\label{sec:additional_qualitative}

\begin{figure*}[h]
    \centering
    \includegraphics[width=\linewidth]{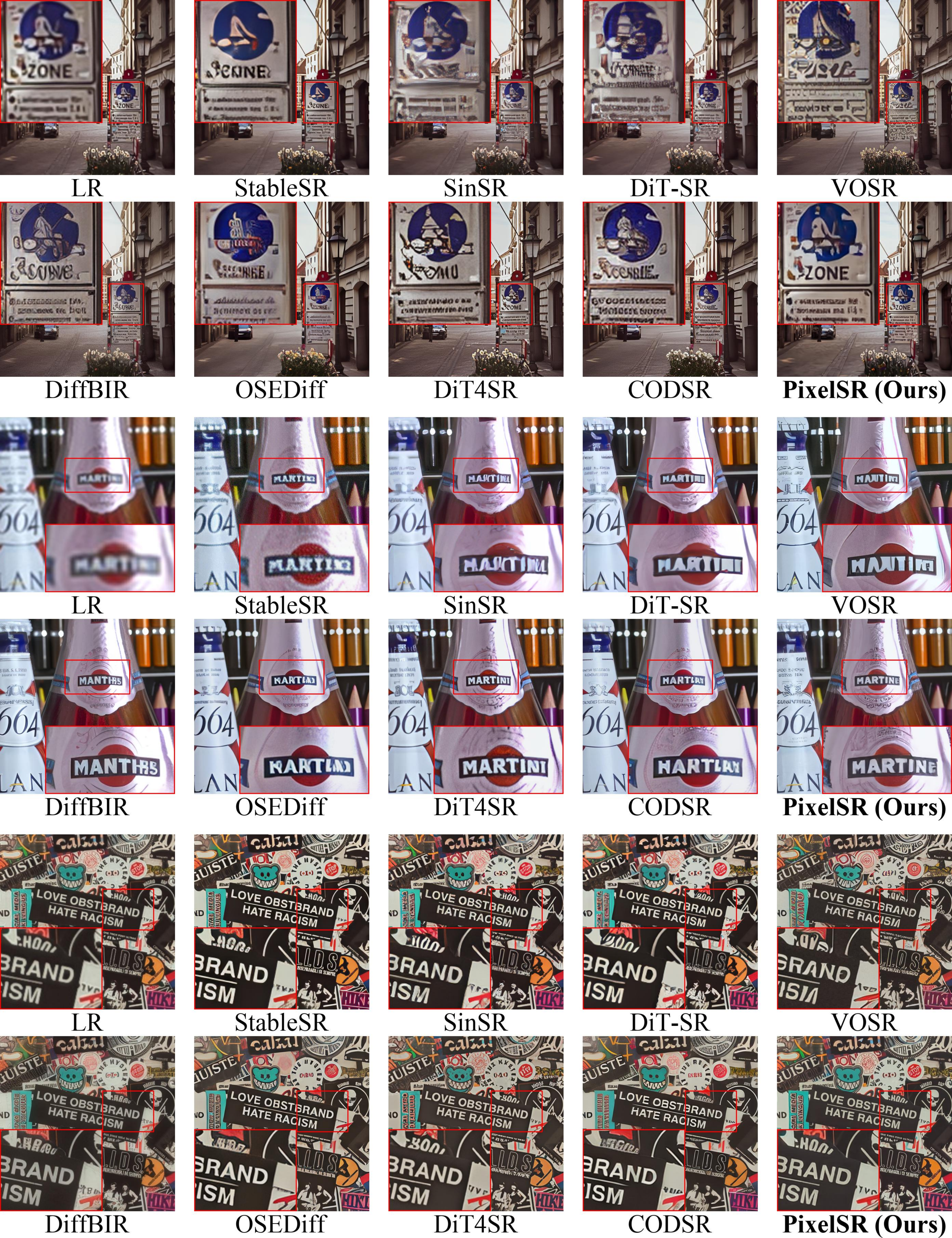}
    \caption{\textbf{Qualitative comparison on challenging real-world examples.} 
    PixelSR better preserves the original text layout and character structures while recovering fine-grained image details, whereas several competing methods exhibit distorted characters, irregular textures, or over-smoothed results. Best viewed zoomed in.
    }
    \label{fig:supp1}
\end{figure*}

\begin{figure*}[t]
    \centering
    \includegraphics[width=\linewidth]{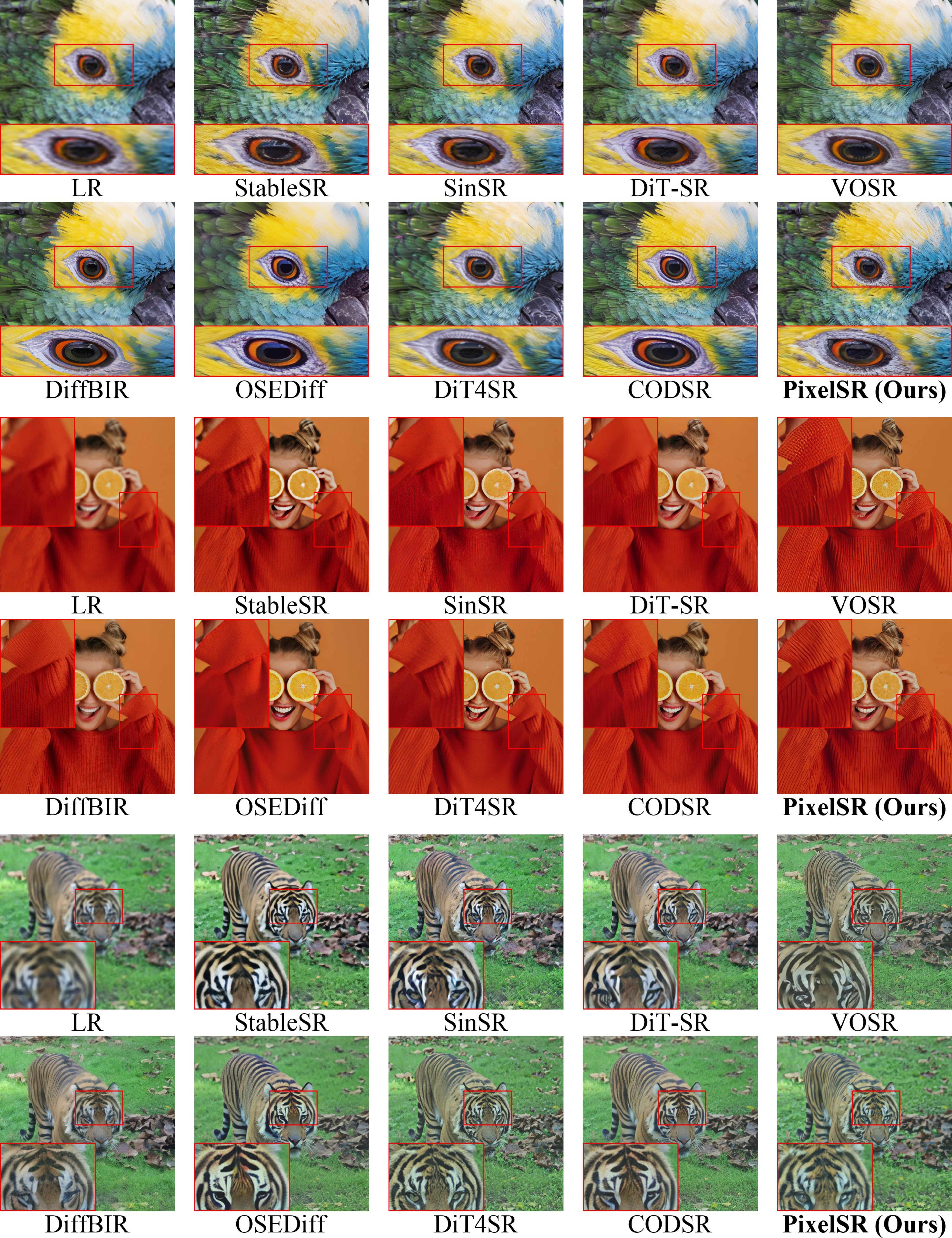}
    \caption{\textbf{Qualitative comparison on challenging real-world examples.} 
    PixelSR better preserves high-frequency textures and fine structural details, whereas competing methods often produce over-smoothed surfaces, irregular patterns, or artificial artifacts. Best viewed zoomed in.
    }
    \label{fig:supp2}
\end{figure*}

Figure~\ref{fig:supp1} and Figure~\ref{fig:supp2} present additional qualitative comparisons on challenging real-world cases.
Competing methods often suffer from over-smoothed textures, distorted structures, inaccurate text restoration, or unrealistic hallucinations.
In contrast, our PixelSR preserves consistent structure while recovering clearer text and finer, more natural textures. These visual results further corroborate the quantitative evidence that our single-stage, end-to-end PixelSR better reconciles structural fidelity and perceptual realism.


\end{document}